%% file: main.tex
\PassOptionsToPackage{table}{xcolor}
\PassOptionsToPackage{capitalize}{cleveref}
\documentclass[10pt,letterpaper]{alaya}
\usepackage[utf8]{inputenc}
\usepackage[sc]{mathpazo}
\usepackage{amsmath,amssymb}
\usepackage[authoryear,round]{natbib}
\usepackage{url}
\usepackage{float}
\let\cite\citep
\newcommand{\Description}[1]{} 
\newcommand{\dataset}{MUGEN\xspace}
\newcommand{\method}{Wan360\xspace}
\definecolor{tableblue}{rgb}{.9,.9,1}
\hypersetup{pdftitle={MUGEN: Interactive Panoramic World Exploration via Camera Control},pdfauthor={Jiaming Tan, Zhen Li, Shuwei Shi, Minggui Teng, Siqi Yang, Yuwei Wu, Bo Zheng, Chuanhao Li, Kaipeng Zhang}}
\RequirePackage{xspace}
\makeatletter
\DeclareRobustCommand\onedot{\futurelet\@let@token\@onedot}
\def\@onedot{\ifx\@let@token.\else.\null\fi\xspace}

\makeatother

\title{MUGEN: Interactive Panoramic World Exploration via Camera Control}
\author[1,2,*]{Jiaming Tan}
\author[1]{Zhen Li}
\author[1]{Shuwei Shi}
\author[1]{Minggui Teng}
\author[1]{Siqi Yang}
\author[2,\dagger]{Yuwei Wu}
\author[1]{Bo Zheng}
\author[1,\dagger]{Chuanhao Li}
\author[1,3,\dagger]{Kaipeng Zhang}
\affiliation[1]{Alaya Lab}
\affiliation[2]{Beijing Institute of Technology}
\affiliation[3]{Shanghai Innovation Institute}
\abstract{

\input{sec/0_abstract}}
\github{\url{https://alaya-lab.github.io/MUGEN}}
\metadata[Code \& Data]{\url{https://github.com/AlayaLab/MUGEN}}
\correspondence{{\footnotesize wuyuwei@bit.edu.cn, chuanhao.li@shanda.com, kaipeng.zhang@shanda.com}}
\date{September 29, 2026}
\begin{document}
\maketitle
\begingroup
\renewcommand{\thefootnote}{\fnsymbol{footnote}}
\footnotetext[1]{Work done during an internship in Alaya Lab.\quad\textsuperscript{\textdagger}Corresponding authors.}
\endgroup
\begin{figure}[H]
    \centering
    \includegraphics[width=\textwidth]{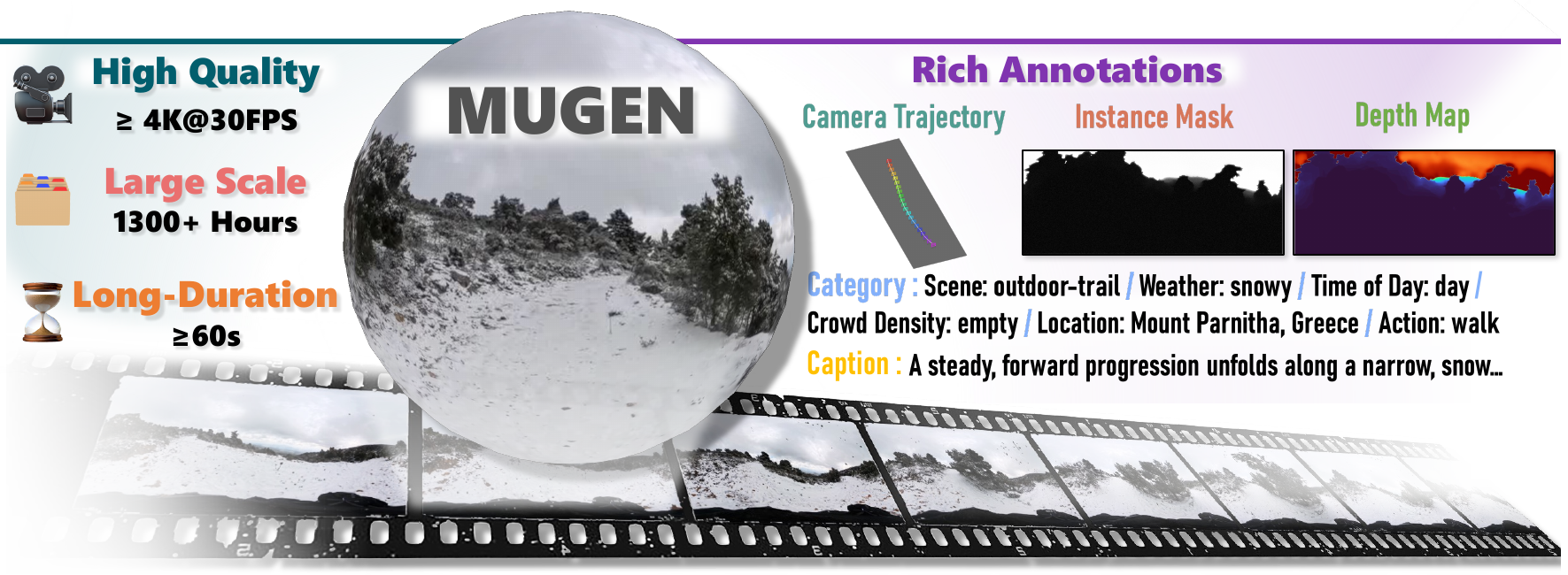}
    \caption{\small MUGEN dataset overview. We introduce a large-scale panoramic video dataset featuring high-quality, long-duration 360° videos totaling 1,300+ hours at the resolution of at least 4K, paired with rich multi-level annotations including camera trajectories, instance masks, depth maps, natural language captions, and structured semantic labels. }
    \Description{An overview graphic centered on a snowy panoramic video from MUGEN. Labels on the left state at least 4K resolution at 30 frames per second, more than 1,300 hours of video, and clips at least 60 seconds long. Examples on the right show a camera trajectory, an instance mask, a depth map, categorical labels for scene, weather, time of day, crowd density, location, and action, and a natural-language caption.}
    \label{fig:overview}
\end{figure}
\clearpage
\input{sec/1_intro}
\input{sec/2_related}

\input{sec/3_dataset}
\input{sec/4_method}
\input{sec/5_experiment}

\FloatBarrier

\input{sec/6_conclusion}
\FloatBarrier
\bibliographystyle{plainnat}
\bibliography{references}

\end{document}

%% file: sec/0_abstract.tex
Interactive panoramic video generation aims to synthesize immersive 360\textdegree{} videos that remain visually coherent while following user-specified camera trajectories during exploration.
However, progress is limited by a coupled data-and-model gap: existing panoramic video datasets are often short, weakly annotated, or lack camera trajectories, while existing camera-controlled video generation models are designed for perspective videos and do not directly support panoramic geometry.
In this paper, we introduce \textbf{MUGEN} and \textbf{Wan360} to address these limitations.
\textbf{MUGEN} is a large-scale real-world panoramic video dataset tailored to interactive 360\textdegree{} world exploration, comprising over 1,300 hours of at least 4K panoramic videos with rich semantic and geometric annotations.
Built on MUGEN, we further present \textbf{Wan360}, a camera-controllable interactive panoramic video generation model.
Panoramic videos are commonly represented by EquiRectangular Projection (ERP), which unfolds a spherical 360\textdegree{} view into a rectangular frame with cyclic longitude seams and pole distortions.
To this end, Wan360 introduces three parameter-free ERP-aware components: periodic longitude RoPE for seam-consistent positional encoding, ERP-aware padding for reducing boundary artifacts, and random roll yaw for consistent learning.
For camera control, Wan360 uses a panoramic Pl\"ucker embedding that represents camera motion with ERP rays rather than perspective pinhole rays.
Experiments show that MUGEN serves as a data foundation for panoramic world exploration, and that Wan360 enables high-quality, temporally coherent, camera-controllable 360\textdegree{} video generation.

%% file: sec/1_intro.tex
\section{Introduction}
\label{sec:intro}

Panoramic videos provide complete 360\textdegree{} visual observations of real-world environments, making them a natural medium for immersive world exploration in virtual reality, simulation, and embodied AI~\cite{yin2025panoworld, chiariotti2021survey,anderson2018vision}.
For such exploration to be truly interactive, a generative system must satisfy three requirements simultaneously: it should synthesize high-fidelity panoramic content, maintain temporal and spherical consistency over time, and follow user-specified camera trajectories during generation.
Recent advances in panoramic video generation have improved the visual quality of 360\textdegree{} videos~\cite{wang2024360dvd,liu2025dynamicscaler,tan2024imagine360,xie2025videopanda,fang2025viewpointpanoramicvideogeneration,xia2025panowan,yin2025panoworld}, but interactive panoramic world exploration remains underdeveloped because both the data and the model architectures are still insufficient for controllable generation in dynamic real-world scenes.

From the data perspective, existing panoramic video datasets are not designed for world exploration.
Generation-oriented datasets \ ~\cite{wang2024360dvd,xia2025panowan} mainly provide short captioned clips, which are useful for text-conditioned panoramic synthesis but do not provide explicit camera-trajectory annotations.
Perception-oriented datasets~\cite{huang2023360vot,xu2025360vots,yan2024panovos,zhang2025leader360v} include masks, boxes, or tracking annotations, but they are typically limited in scale and duration and do not target camera control video generation.
As a result, prior datasets rarely combine high-resolution real-world panoramic videos, minute-level temporal extent, semantic annotations, and geometric annotations such as camera trajectories.
Lacking of such a dataset makes it difficult to train and evaluate models that can generate long, coherent, and controllable panoramic videos.
To address this data gap, we introduce \textbf{MUGEN}, shown in \cref{fig:overview}, a large-scale real-world panoramic video dataset tailored to interactive 360\textdegree{} world exploration.
MUGEN contains over 1,300 hours of high-fidelity panoramic videos at a resolution of at least 4K, with source videos longer than one minute and standardized one-minute clips for annotation and training.
Each clip is paired with rich semantic annotations, including captions, scene categories, actions, weather, crowd density, and locations, as well as geometric annotations including camera trajectories, instance masks, and depth.
To construct MUGEN, we design an automatic pipeline that collects panoramic videos from curated YouTube channels, performs format and content filtering, segments videos into minute-level clips, and annotates them using large multimodal models and geometry estimation tools.
We further construct MUGEN-HQ, a 300-hour high-quality subset selected by quality and diversity, to support efficient model training and evaluation.

From the model perspective, directly extending existing video generation models to interactive panoramic generation is also non-trivial.
Panoramic videos are usually represented in equirectangular projection (ERP), whose geometry differs fundamentally from ordinary perspective frames: longitude is periodic, the left and right image boundaries correspond to the same seam, and the top and bottom rows collapse near the poles.
Treating ERP frames as standard rasters can therefore introduce seam artifacts, pole artifacts, and spherical inconsistencies.
To address this model gap, we propose \textbf{Wan360}, a camera-controllable panoramic image-to-video model trained on MUGEN-HQ.
Wan360 is fine-tuned from the perspective camera-control baseline Wan2.2-Fun-5B-Control-Camera~\cite{wan2025} and adapts it to ERP panoramas through three parameter-free ERP-aware components.
\textbf{Periodic longitude RoPE} enforces cyclic positional encoding along the horizontal longitude dimension, \textbf{ERP-aware padding} reduces seam artifacts in VAE encoding and decoding, and \textbf{random roll yaw} exposes the model to yaw-equivalent panoramic observations while preserving the corresponding camera trajectories.
For camera control, Wan360 uses a panoramic Pl\"ucker embedding~\cite{ji2025campvg}, replacing perspective pinhole rays with ERP rays to enable trajectory conditioning without conventional camera intrinsics.
These components allow Wan360 to reuse the pretrained video generation baseline while avoiding additional trainable modules.

Experiments demonstrate the effectiveness of both MUGEN and Wan360.
We evaluate the annotation quality of MUGEN and show that its semantic and geometric annotations provide reliable data support for interactive panoramic world exploration.
We further compare Wan360 with existing panoramic video generation methods and conduct ablations on its ERP-aware components.
The results show that Wan360 achieves strong panoramic video quality, temporal coherence, and camera-trajectory controllability, indicating that the improvements come from both the scale and quality of MUGEN and the proposed ERP-aware components.

To sum up, our contributions are as follows:
(i) We propose \textbf{MUGEN}, a large-scale, long-duration, high-quality real-world panoramic video dataset for interactive panoramic world exploration, containing over 1,300 hours of videos with rich semantic and geometric annotations.
(ii) We present \textbf{Wan360}, a camera-controllable panoramic video generation model that adapts to ERP geometry via three parameter-free ERP-aware components and a panoramic Pl\"ucker embedding.
(iii) Extensive experiments validate the annotation quality of MUGEN and the generation quality of Wan360.

%% file: sec/2_related.tex
\section{Related Work}
\label{sec:rela}

\subsection{Panoramic Video Datasets}

Existing panoramic video datasets mainly support either generation or perception.
Generation-oriented datasets such as WEB360~\cite{wang2024360dvd} and PANOVID~\cite{xia2025panowan} provide captioned panoramic clips for text-conditioned synthesis, but their annotations are primarily semantic and lack explicit camera trajectories. 360-1M~\cite{wallingford2024image} targets static novel view synthesis from frame pairs, PanFlow~\cite{zhang2026panflow} curates a motion-rich panoramic dataset with frame-level pose and flow annotations tailored to optical-flow-conditioned motion control, whereas MUGEN provides trajectories and richer annotations, better suiting panoramic world exploration. 
Perception-oriented datasets such as 360VOT/360VOTS~\cite{huang2023360vot,xu2025360vots}, PanoVOS~\cite{yan2024panovos}, and Leader360V~\cite{zhang2025leader360v} provide tracking or segmentation annotations, but are not designed for world exploration.
In contrast, MUGEN is the first dataset with large-scale real-world 360\textdegree{} videos, minute-level duration, high visual fidelity, semantic labels, and geometric annotations including camera trajectories, depth, and instance masks to support interactive world exploration.

\subsection{Panoramic Video Generation}

Recent panoramic video generation methods adapt general video priors to ERP geometry using panorama-specific adapters~\cite{wang2024360dvd}, scalable generation~\cite{liu2025dynamicscaler}, spherical or panorama-aware representations~\cite{park2025spherediff,zhang2025panodit,xie2025videopanda,xia2025panowan}, and perspective-to-panorama lifting~\cite{tan2024imagine360,lu2025genexgeneratingexplorableworld,fang2025viewpointpanoramicvideogeneration,li20244k4dgenpanoramic4dgeneration,li2026cubecomposer}.
These works improve visual fidelity and scene coverage, but ERP panoramic video generation still requires geometry-aware mechanisms to handle longitude periodicity, boundary artifacts, and yaw-equivalent observations.
Wan360 achieves higher-quality panoramic video generation through three parameter-free ERP-aware components that address these geometric properties.
These components share related geometric intuitions with several prior works, but differ in how they are formulated and applied. PanoSplatt3R~\cite{ren2025panosplatt3r} approximates ERP periodicity through head-wise rolled linear RoPE coordinates. In contrast, our Periodic Longitude RoPE is exactly cyclic for every head, leading to better continuity at the seam. PAR~\cite{wang2026conditional} applies circular padding to panoramic image generation, whereas we extend it to the 3D video VAE for video generation. PanoDiffusion~\cite{wu2023panodiffusion} rotates the panorama during training, our Random Roll Yaw also rotates the camera trajectory simultaneously, thereby enhancing the diversity of camera trajectories during training.

\subsection{Camera-Controlled Video Generation}

Camera-controlled video generation typically injects control signals into the model in two ways: by converting camera motion into Pl\"ucker embeddings~\cite{he2024cameractrl}, or by adding explicit discrete control cues into the text prompts~\cite{wan2025}.
Existing methods mainly focus on conventional perspective video generation~\cite{alayaworld2026short,alayaworld2026full,alayaworld2026v11,he2025matrixgame2,sun2025worldplay,zhu2026sanawm,robbyant2026lingbotworld}.  
Although PanoWorld-X~\cite{yin2025panoworld}, CamPVG~\cite{ji2025campvg}, and OmniRoam~\cite{liu2026omniroam} enable camera-controlled interactive panoramic video generation, these methods operate primarily in static or quasi-static environments.
In contrast, Wan360 targets dynamic real-world panoramic world exploration, emphasizing controllable generation under natural, diverse, and unconstrained conditions.

%% file: sec/3_dataset.tex
\section{\dataset Curation}
\label{sec:dataset}

\begin{figure*}[t]
  \centering
  \includegraphics[width=1.0\linewidth]{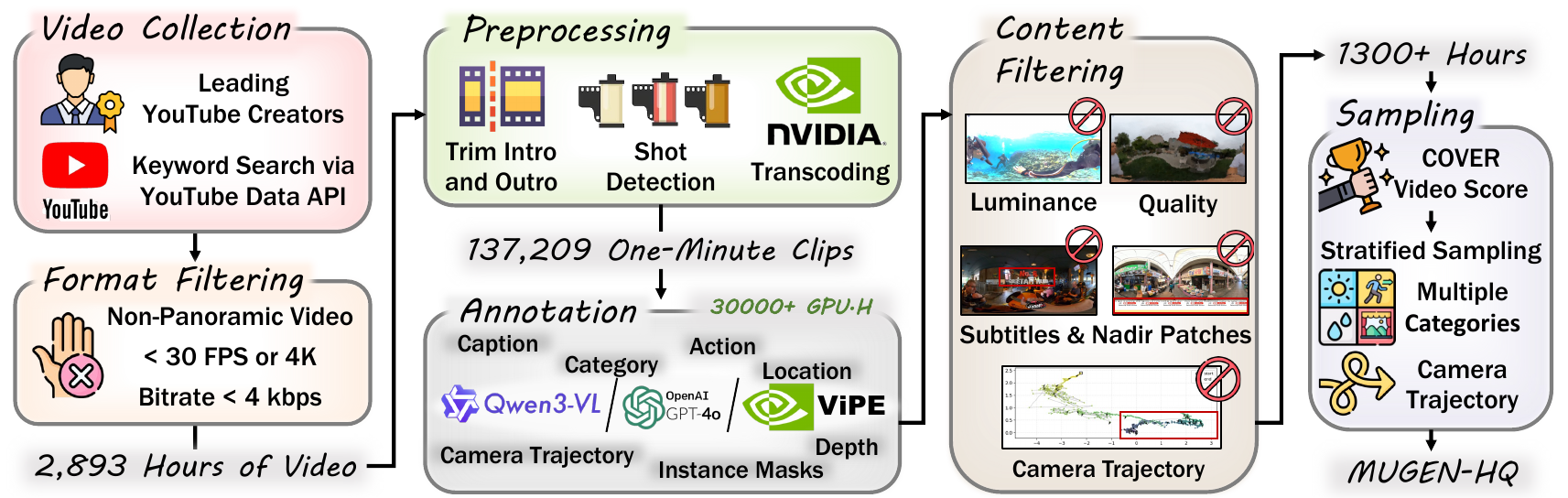}
  
  \caption{Data curation pipeline for MUGEN. We collect 2,893 hours of panoramic source videos from YouTube, preprocess continuous footage into consecutive one-minute clips for scalable annotation, annotate each clip with semantic and geometric information using Qwen3-VL~\cite{yang2025qwen3}, GPT-4o~\cite{hurst2024gpt}, and ViPE~\cite{huang2025vipe}, filter clips by visual quality, overlays, and trajectory validity, and finally sample MUGEN-HQ for model training. The pipeline yields MUGEN (1,318 hours) and MUGEN-HQ (300 hours).}
  \Description{A left-to-right flowchart of MUGEN data curation. YouTube videos from leading creators and keyword search first pass format filters for panoramic format, frame rate, resolution, and bitrate. The resulting 2,893 hours are trimmed, split by shot, and transcoded into 137,209 one-minute clips. Qwen3-VL, GPT-4o, and ViPE annotate captions, categories, actions, locations, camera trajectories, instance masks, and depth. Content and trajectory filters produce more than 1,300 hours, from which stratified sampling by video score, category, and camera trajectory forms MUGEN-HQ.}
  \label{fig:pipeline}
\end{figure*}

To address the lack of large-scale, high-quality, richly annotated panoramic video datasets tailored to world exploration, we introduce MUGEN.
\cref{fig:pipeline} presents the data curation pipeline for our MUGEN dataset.
We begin by collecting panoramic videos from YouTube, focusing on diverse real-world environments.
To ensure high data fidelity, we apply strict format filtering based on resolution, bitrate, and frame rate.
The filtered videos are then preprocessed into minute-level clips through intro/outro trimming, shot boundary detection, and standardized transcoding.
Each clip is annotated with comprehensive semantic and geometric information using multimodal models, followed by content-based quality filtering to produce the final MUGEN dataset.
Furthermore, we construct MUGEN-HQ by sampling a high-quality subset with the most reliable and diverse annotations.
MUGEN also provides panoramic video supervision for AlayaVista~\cite{alayavista2026}, which studies streaming world modeling from panoramic states to perspective observations.

\subsection{Data Acquisition and Preprocessing}

We use YouTube as the primary source for diverse real-world panoramic videos.
We manually curate leading panoramic video channels and search the YouTube Data API with queries such as ``360 tour'', then review the retrieved results to exclude non-real-world content such as simulation or rendered scenes.
This stage yields 9,544 candidate source videos.

We further query video metadata and apply format filtering to retain high-fidelity panoramic videos.
We remove non-panoramic videos and videos with frame rates below 30 FPS, resolutions below 4K, or bitrates below 4 kbps.
For the remaining videos, we download the highest available resolution and best-quality audio, resulting in 2,893 hours of raw panoramic footage.
All retained source videos are longer than one minute.

For preprocessing, we first verify file integrity and discard corrupted videos.
We trim the first and last minute of each source video to remove common intro and outro segments.
Because online videos are often edited, we use TransNetV2~\cite{soucek2020transnetv2} to detect shot boundaries and keep only temporally continuous segments.
Following Sekai~\cite{li2025sekai}, we accelerate shot-boundary processing with PyNvVideoCodec and CVCUDA.
Each continuous segment is partitioned into consecutive non-overlapping 60-second clips, which makes large-scale annotation tractable and provides a consistent training unit.
The clips are transcoded to H.265 at the original resolution and frame rate, with audio re-encoded to AAC at 48 kHz.
This process produces 137,209 one-minute clips totaling approximately 2,287 hours before content filtering.

\subsection{Annotation}

MUGEN provides two groups of annotations: semantic annotations for content generation and geometric annotations for interactive camera control.

\subsubsection{Semantic Annotations}
Because ERP panoramas are difficult for vision-language models to parse directly, we project each ERP clip into six cube faces with a 90\textdegree{} $\times$ 90\textdegree{} field of view and attach explicit face labels.
We use Qwen3-VL-235B-A22B-Instruct~\cite{yang2025qwen3} to generate detailed clip-level narratives of approximately 250 words, describing scene layout, visible objects, events, and camera motion.
To support different training and evaluation settings, we also provide event-agnostic captions and interval captions at 5s/10s granularity.
We further prompt Qwen3-VL with YouTube metadata to predict structured attributes, including weather, time of day, crowd density, scene category, and action.
Following SpatialVID~\cite{wang2025spatialvid}, we refine coarse indoor/outdoor labels into fine-grained scene categories and introduce action labels tailored to world exploration.
For geographic information, we use GPT-4o~\cite{hurst2024gpt} to extract location tags from video metadata; clips with ambiguous or missing locations are marked as \emph{unspecified}.

\subsubsection{Geometric Annotations}
For camera-controllable generation, each clip requires trajectory annotations.
We compare VGGT~\cite{wang2025vggt}, MegaSAM~\cite{li2025megasam}, and ViPE~\cite{huang2025vipe} on sampled panoramic clips, and choose ViPE because it directly supports panoramic video input and produces more coherent camera trajectories in our setting.
We use ViPE to estimate camera trajectories, instance masks, and depth for each clip.
Generating geometric annotations for the full dataset requires over 30,000 GPU hours.

\subsection{Quality Filtering}

Raw web videos inevitably contain poor lighting, blur, overlays, stitching artifacts, or unreliable trajectories.
We therefore apply four content filters before forming MUGEN.
\subsubsection{Luminance}
We compute the mean luma value of each clip and discard clips outside the range [40, 215] to remove underexposed or overexposed content.

\subsubsection{Visual Quality}
We score clips with COVER~\cite{cover2024cpvrws} and remove clips below 0.7, which typically contain severe blur, distortion, or compression artifacts.

\subsubsection{Overlays}
We project each ERP clip into six perspective views and use Qwen3-VL~\cite{yang2025qwen3} to flag subtitles, watermarks, and nadir patches.

\subsubsection{Trajectory Validity}
We filter clips whose estimated camera trajectories contain chaotic paths, temporal discontinuities, or abnormal rotations.
After filtering, MUGEN contains 1,318 hours of high-quality one-minute clips from 6,446 unique source videos.

\subsection{MUGEN-HQ Sampling}

Although MUGEN is suitable for large-scale training and evaluation, model development often requires a smaller subset with high visual quality and balanced coverage.
We therefore construct MUGEN-HQ, a 300-hour subset used for training Wan360.
We first rank clips by COVER score and retain the top 70\% as a quality-filtered pool.
We then perform stratified sampling across scene categories, camera-motion patterns, action types, and weather conditions.
This strategy preserves high visual quality while improving diversity in both scene content and controllable camera motion.
MUGEN-HQ contains clips from 4,023 unique source videos.

\subsection{Dataset Analysis}

\begin{figure}
  \centering
  \scriptsize
  \setlength{\tabcolsep}{3pt}
  \renewcommand{\arraystretch}{1.05}
  \begin{minipage}{0.49\linewidth}
    \centering
    \includegraphics[width=\linewidth]{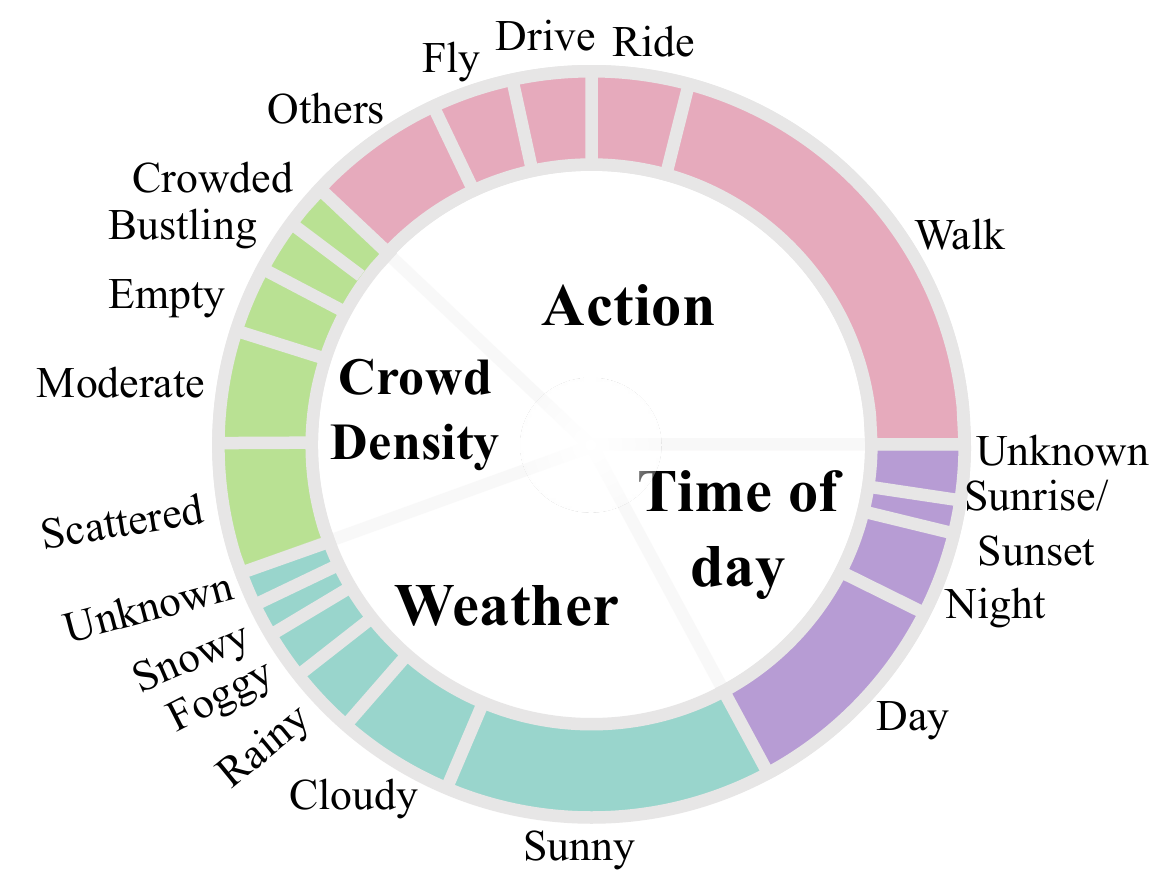}
  \end{minipage}\hfill
  \begin{minipage}{0.49\linewidth}
    \centering
    \begin{tabular}{l r}
      \toprule
      \textbf{Statistic} & \textbf{Value} \\
      \midrule
      \multicolumn{2}{l}{\textbf{Translation}} \\
      \quad Avg.\ Path Length & 18.16 $\pm$ 14.74 \\
      \quad Avg.\ Max Displacement & 15.02 $\pm$ 13.42 \\
      \quad Avg.\ Speed (unit/s) & 0.30 $\pm$ 0.25 \\
      \midrule
      \multicolumn{2}{l}{\textbf{Rotation}} \\
      \quad Avg.\ Total Rotation ($^\circ$) & 128.0 $\pm$ 192.3 \\
      \quad Avg.\ Net Rotation ($^\circ$) & 30.6 $\pm$ 40.9 \\
      \quad Avg.\ Angular Speed ($^\circ$/s) & 2.1 $\pm$ 3.2 \\
      \midrule
      \multicolumn{2}{l}{\textbf{Dominant Motion Direction}} \\
      \quad Forward / Backward & 44.3\% / 17.3\% \\
      \quad Left / Right & 17.3\% / 18.2\% \\
      \quad Up / Down & 0.5\% / 0.4\% \\
      \quad Static & 2.1\% \\
      \bottomrule
    \end{tabular}
  \end{minipage}
  \caption{Statistics of semantic attributes and camera trajectories in MUGEN. Left: semantic attribute distributions. Right: camera trajectory statistics.}
  \Description{The left radial chart groups semantic attributes into action, time of day, weather, and crowd density. Walking is the largest action category; day and sunny are the largest time and weather categories; and the crowd labels range from empty to crowded. The right table reports an average path length of 18.16, average maximum displacement of 15.02, average total rotation of 128.0 degrees, and average net rotation of 30.6 degrees. Forward motion is most common at 44.3 percent, followed by right, backward, and left motion at roughly 17 to 18 percent each; vertical and static motions are rare.}
  \label{fig:dataset_stats}
\end{figure}

MUGEN covers diverse semantic and motion conditions for panoramic world exploration.
As shown in \cref{fig:dataset_stats}, the dataset spans multiple weather types, times of day, crowd-density levels, scene categories, and action types.
The trajectory statistics show substantial variation in translation length, rotation magnitude, angular speed, and dominant motion direction.
These properties provide the semantic diversity and trajectory diversity needed to train and evaluate camera-controllable panoramic generation models.
We will release metadata and processing scripts to support transparent reuse.

%% file: sec/4_method.tex
\section{\method Model}
\label{sec:model}

\begin{figure*}[t]
  \centering
  \includegraphics[width=1.0\textwidth]{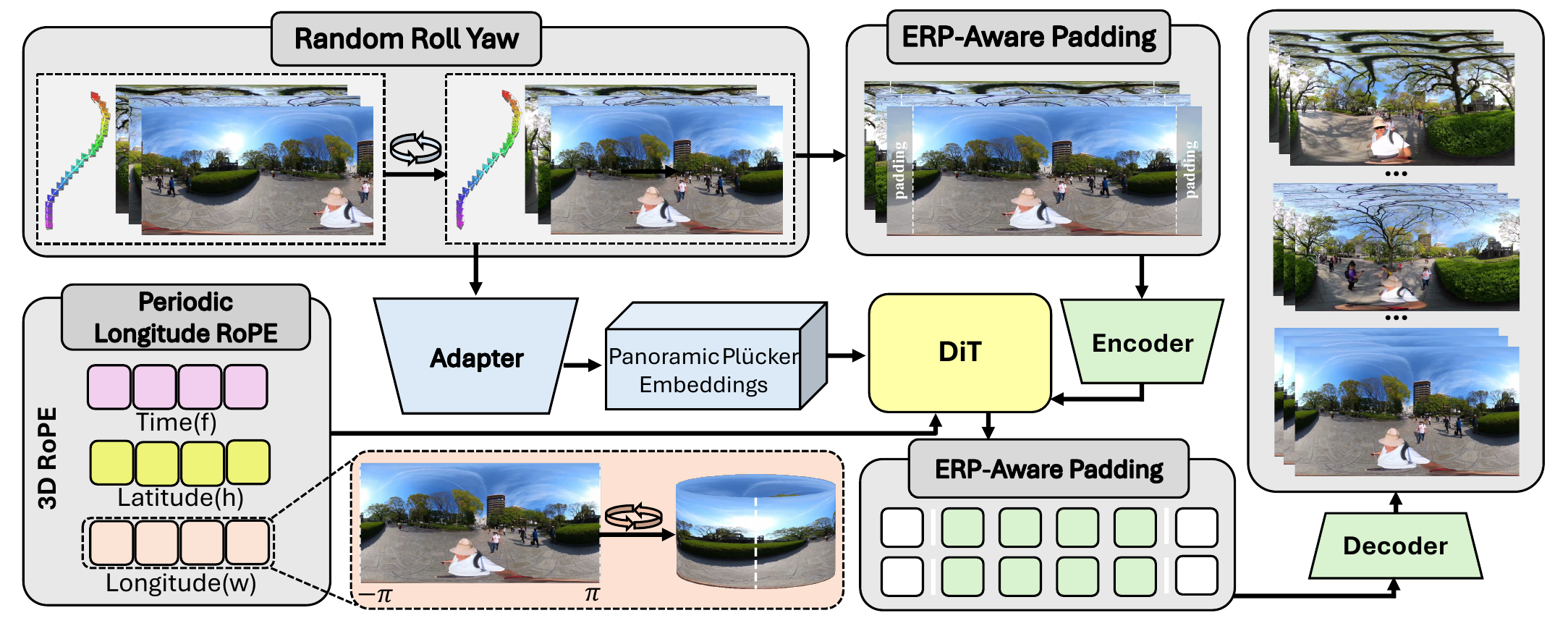}
  \caption{Architecture of \method. Wan360 is a camera-controllable interactive panoramic video generation model towards world exploration conditioned on an input panorama, a text prompt, and a target camera trajectory. To adapt the perspective baseline to panoramic geometry, Wan360 introduces three parameter-free ERP-aware components: periodic longitude RoPE for cyclic longitude encoding, ERP-aware padding for seam-continuous VAE features, and random roll yaw for yaw-equivalent trajectory augmentation. A panoramic Pl\"ucker embedding converts ERP pixels into spherical rays and injects trajectory conditions through the pretrained control adapter.}
  \Description{A block diagram of the Wan360 architecture. The input panorama and camera trajectory first receive a matched random roll-yaw transformation. The trajectory passes through an adapter and panoramic Pluecker embedding into a diffusion transformer, while an encoder supplies panorama features using ERP-aware boundary padding. Periodic longitude rotary position encoding augments the transformer's spatial coordinates. The transformer output passes through ERP-aware padding and a decoder to produce a sequence of panoramic video frames.}
  \label{fig:arch}
\end{figure*}

We present \textbf{Wan360}, a panoramic image-to-video (I2V) generation model that generates ERP videos from an input panorama, a text prompt, and a user-specified camera trajectory.
Wan360 is fine-tuned from a perspective camera-control video generation baseline to preserve its strong generation and control priors, but this baseline cannot be directly applied to panoramas.
ERP videos contain cyclic longitude seams and pole singularities, and panoramic cameras do not follow the perspective pinhole model.
Naive fine-tuning would therefore introduce seam artifacts, spherical inconsistencies, and mismatched camera-control signals.

\subsection{Component Overview}
\Cref{fig:arch} shows the overall architecture.
Wan360 addresses the ERP geometry mismatch with three components that add no trainable parameters.
\textit{Periodic Longitude RoPE} (\cref{subsec:rope}) makes longitude positional encoding seam-consistent, \textit{ERP-Aware Padding} (\cref{subsec:pad}) reduces boundary artifacts in VAE encoding and decoding, and \textit{Random Roll Yaw} (\cref{subsec:yaw}) creates yaw-equivalent panorama-trajectory training pairs.
For camera control, a \textit{Panoramic Pl\"ucker Embedding} (\cref{subsec:ppe}) replaces the baseline's pinhole ray-condition generator with an ERP ray generator, enabling intrinsics-free trajectory conditioning.

\subsection{Periodic Longitude RoPE}
\label{subsec:rope}

Perspective video baselines treat the horizontal image axis as a bounded coordinate, but ERP longitude is periodic: the left and right image boundaries represent adjacent directions on the sphere.
Using the original positional encoding therefore introduces an artificial discontinuity at the panorama seam.
Periodic Longitude RoPE replaces only the longitude part of the positional encoding with a cyclic formulation, while keeping the temporal and latitude parts unchanged for compatibility with the pretrained baseline.
For a token at horizontal index $u$ in a width-$W$ latent grid, we map it to longitude
\begin{equation}
    \theta_u = 2\pi \frac{u+1/2}{W} - \pi,
    \qquad
    \rho_k(u)=\big(\cos(k\theta_u), \sin(k\theta_u)\big),
\end{equation}
where $k$ denotes the RoPE frequency.
Because $\rho_k(u+W)=\rho_k(u)$, the positional code is continuous across the ERP seam.
This makes tokens near the two horizontal boundaries geometrically consistent and reduces seam-related artifacts during fine-tuning.

\subsection{ERP-Aware Padding}
\label{subsec:pad}

The VAE in the baseline uses local convolutions, whose default boundary handling assumes that image edges have no continuation.
For ERP panoramas this assumption is incorrect, because pixels at the two horizontal borders are neighbors on the sphere.
ERP-Aware Padding exposes this cross-boundary neighborhood by using longitude-cyclic padding during VAE encoding and decoding.
For any VAE feature map $z$, horizontal padding is defined as
\begin{equation}
    \tilde{z}_{t,v,u} = z_{t,v,\,u \bmod W},
\end{equation}
so convolutional neighborhoods that cross the left or right border are sampled from the opposite side of the panorama.
This prevents the autoencoder from injecting artificial boundary artifacts into latent features and decoded frames.

\subsection{Random Roll Yaw}
\label{subsec:yaw}

ERP panoramas have no canonical yaw orientation: horizontally rolling a panorama corresponds to rotating the world around the vertical axis.
If the model only sees the yaw distribution in the collected videos, it may overfit to dataset-specific orientations and become less robust under user-specified trajectories.
Random Roll Yaw augments each training sample with a yaw-equivalent version by rolling the ERP video and applying the same yaw rotation to the camera trajectory.
Given a horizontal roll $s$, the augmented video and camera-to-world pose are
\begin{equation}
    I'_t(u,v)=I_t((u-s)\bmod W,v),
    \qquad
    T'_t = R_y(2\pi s/W)\,T_t .
\end{equation}
This keeps the video and control signal geometrically aligned while exposing the model to diverse trajectory orientations for the same scene.

\subsection{Panoramic Pl\"ucker Embedding}
\label{subsec:ppe}

Perspective camera-control methods usually describe camera motion through pinhole rays parameterized by camera intrinsics.
This representation is not suitable for ERP panoramas, where each pixel corresponds to a direction on the viewing sphere rather than a ray through a perspective image plane.
We therefore formulate the control signal as a panoramic Pl\"ucker field: ERP pixels are converted to spherical rays and then transformed by the target camera poses.
For an ERP pixel $(u,v)$, we first compute its longitude and latitude,
\begin{equation}
    \lambda_u = 2\pi \frac{u+1/2}{W} - \pi,
    \qquad
    \phi_v = \frac{\pi}{2} - \pi \frac{v+1/2}{H},
\end{equation}
and obtain the camera-space ray direction
\begin{equation}
    d_{\mathrm{cam}}(u,v)=
    \begin{bmatrix}
    \cos\phi_v \sin\lambda_u \\
    \sin\phi_v \\
    \cos\phi_v \cos\lambda_u
    \end{bmatrix}.
\end{equation}
With camera pose $(R_t,o_t)$, the panoramic Pl\"ucker condition is
\begin{equation}
    d_t = R_t d_{\mathrm{cam}},
    \qquad
    m_t = o_t \times d_t,
    \qquad
    L_t(u,v)=\big[d_t; m_t\big].
\end{equation}
This provides trajectory conditioning that is compatible with panoramic geometry while reusing the baseline's ray-based control interface.

%% file: sec/5_experiment.tex
\section{Experiments}
\label{experiment}

\begin{table*}[!t]
  \centering
  \small
  \setlength{\tabcolsep}{4.0pt}
  \renewcommand{\arraystretch}{1.15}
  \caption{Experimental results on generated video quality and camera control accuracy. (I) Ablation of the three ERP-aware components in \method; (II) disentangled evaluation of model and data; (III) comparison with other panoramic video generation methods. $\uparrow$ / $\downarrow$ indicates whether higher / lower is better.}
  \resizebox{\textwidth}{!}{%
  \begin{tabular}{lcccccccc}
    \toprule
    \textbf{Model} &
    \textbf{FVD} $\downarrow$ &
    \textbf{SSIM} $\uparrow$ &
    \textbf{LPIPS} $\downarrow$ &
    \textbf{Consistency} $\uparrow$ &
    \textbf{Quality} $\uparrow$ &
    \textbf{Dynamic} $\uparrow$ &
    \textbf{PSNR} $\uparrow$ &
    \textbf{TransErr} $\downarrow$ \\
    \midrule

    \multicolumn{9}{l}{\textbf{(I) Ablation of the three ERP-aware components.}} \\
    Wan360 (w/o Periodic Longitude RoPE) &
      491.2 & 0.435 & 0.393 & \underline{0.912} & \underline{0.459} & \textbf{0.740} & 14.58 & 0.192 \\
    Wan360 (w/o ERP-Aware Padding) &
      \underline{478.5} & \underline{0.446} & \underline{0.378} & \textbf{0.915} & 0.457 & 0.710 & \underline{14.79} & \underline{0.188} \\
    Wan360 (w/o Random Roll Yaw) &
      486.9 & 0.439 & 0.387 & \underline{0.912} & 0.456 & \underline{0.720} & 14.65 & 0.215 \\
    \rowcolor{tableblue} \textbf{Wan360 (Ours, full)} &
      \textbf{476.6} & \textbf{0.449} & \textbf{0.377} & \textbf{0.915} & \textbf{0.463} & \underline{0.720} & \textbf{14.86} & \textbf{0.184} \\
    \midrule

    \multicolumn{9}{l}{\textbf{(II) Disentangled evaluation of model and data.}} \\
    GenEX~\cite{lu2025genexgeneratingexplorableworld} &
      1359 & \underline{0.376} & 0.617 & 0.763 & 0.304 & \textbf{0.891} & 13.13 & -- \\
    Wan360 (GenEX's data) &
      999.5 & 0.340 & 0.493 & 0.872 & 0.387 & \underline{0.825} & 13.18 & -- \\
    ViewPoint~\cite{fang2025viewpointpanoramicvideogeneration} &
      1676 & 0.297 & 0.707 & 0.861 & 0.332 & 0.650 & 10.85 & -- \\
    Wan360 (ViewPoint's data) &
      \underline{952.3} & 0.361 & \underline{0.429} & \underline{0.896} & \underline{0.413} & 0.550 & \underline{14.00} & -- \\
    \rowcolor{tableblue} \textbf{Wan360 (Ours)} &
      \textbf{476.6} & \textbf{0.449} & \textbf{0.377} & \textbf{0.915} & \textbf{0.463} & 0.720 & \textbf{14.86} & \textbf{0.184} \\
    \midrule

    \multicolumn{9}{l}{\textbf{(III) Comparison with other methods.}} \\
    4K4DGen~\cite{li20244k4dgenpanoramic4dgeneration} &
      1541 & 0.345 & 0.494 & \textbf{0.951} & 0.421 & 0.275 & 13.02 & -- \\
    DynamicScaler~\cite{liu2025dynamicscaler} &
      1941 & 0.269 & 0.641 & 0.886 & 0.423 & 0.397 & 10.84 & -- \\
    CubeComposer~\cite{li2026cubecomposer} &
      1579 & 0.305 & 0.579 & \underline{0.918} & 0.395 & 0.100 & 11.46 & -- \\
    OmniRoam~\cite{liu2026omniroam} &
      \underline{640.0} & \underline{0.410} & \underline{0.463} & 0.901 & \underline{0.430} & \textbf{0.723} & \underline{13.77} & \underline{0.251} \\
    \rowcolor{tableblue} \textbf{Wan360 (Ours)} &
      \textbf{476.6} & \textbf{0.449} & \textbf{0.377} & 0.915 & \textbf{0.463} & \underline{0.720} & \textbf{14.86} & \textbf{0.184} \\
    \bottomrule
  \end{tabular}%
  }
  \label{tab:tab-vg-all}
\end{table*}

\subsection{Evaluation of Dataset Quality}

We conduct a user study to evaluate the annotation quality of MUGEN from three complementary aspects: long-form text descriptions, structured category labels, and camera trajectories.
We randomly sample 1,000 one-minute clips from MUGEN and recruit 10 volunteers for human verification.
For each aspect, every sampled clip is independently reviewed by all 10 volunteers.
An annotation is considered valid if at least seven volunteers judge it to be correct or usable.

\subsubsection{Text Descriptions}
Volunteers judge whether each long-form description matches the clip content, events, and camera motion; hallucinated or contradictory descriptions are marked as invalid.
94.7\% of the sampled descriptions are judged to be accurate, indicating that the text annotations provide reliable semantic supervision for panoramic video generation.

\subsubsection{Category Labels}
Volunteers judge whether the provided category labels are supported by the video; ambiguous or visually unverifiable labels are treated as invalid.
The final valid ratio reaches 95.3\%, showing that the automatically generated category labels are sufficiently accurate for dataset analysis and controlled training.

\subsubsection{Camera Trajectories}
Volunteers inspect trajectory-overlaid clips and judge whether each trajectory is plausible and aligned with the observed camera motion; obvious drift or wrong-direction cases are marked as unusable.
The resulting usable trajectory ratio is 92.4\%, validating that MUGEN provides reliable camera-control annotations for interactive panoramic video generation.

\subsection{Experimental Setup}

\subsubsection{Training Details}
Wan360 is fine-tuned from the Wan2.2-Fun-5B-Control-Camera~\cite{wan2025} baseline.
We fine-tune it on MUGEN-HQ to generate $161$ frames at $1920{\times}960$ resolution, sampled at $16$\,FPS from the $30$\,FPS source videos.
Training is conducted on $8\times$NVIDIA H200 GPUs for $10$ epochs with a learning rate of $\mathbf{1{\times}10^{-5}}$.

\subsubsection{Evaluation Data}
We randomly select 200 video samples that are excluded from training.
From them, we curate a final test set of 100 videos by stratifying the samples according to camera trajectory patterns and scene categories.

\subsubsection{Metrics}
For reconstruction-oriented quality, we report FVD~\cite{unterthiner2018towards}, SSIM~\cite{wang2004image}, LPIPS~\cite{zhang2018unreasonable}, and PSNR against held-out reference videos.
For perceptual and temporal quality, we use VBench++~\cite{huang2024vbench++}.
For camera control, following CamCtrl~\cite{he2024cameractrl}, we annotate the trajectory of each generated video using ViPE~\cite{huang2025vipe} and compute the translation error (\emph{TransErr}) against the input trajectory.

\subsection{Evaluation of Model Performance}

\subsubsection{Ablation Study}
Block (I) of \cref{tab:tab-vg-all} evaluates the three ERP-aware components in Wan360.
Removing Periodic Longitude RoPE, ERP-aware padding, or random roll yaw weakens the overall balance between video quality and camera control.
The full model achieves the best overall score, indicating that the three components are complementary for fine-tuning a perspective video generation baseline for panoramic video generation.

\subsubsection{Disentangling Model and Data}
Block (II) isolates the contributions of architecture and data.
We train Wan360 using the same training recipe while replacing MUGEN-HQ with data from GenEX~\cite{lu2025genexgeneratingexplorableworld} or ViewPoint~\cite{fang2025viewpointpanoramicvideogeneration}.
Under the same evaluation protocol, Wan360 trained on these alternative datasets improves several metrics over the corresponding original methods, but remains behind Wan360 trained on MUGEN-HQ.
This result suggests that Wan360 contributes meaningful model-side gains, while the scale, diversity, and annotation quality of MUGEN-HQ are essential for the final performance.

\subsubsection{Comparison with Existing Methods}
Block (III) evaluates Wan360 against representative panoramic video generation methods under the same held-out test set.
Compared with prior methods, Wan360 substantially reduces FVD and LPIPS while improving SSIM, PSNR, and video quality, indicating better frame-level fidelity and temporal realism.
Among methods with explicit camera-control evaluation, Wan360 also achieves the lowest TransErr, showing more accurate trajectory following in dynamic real-world panoramas.

\subsection{Qualitative Examples}

\begin{figure}[!t]
  \centering
  \includegraphics[width=\textwidth]{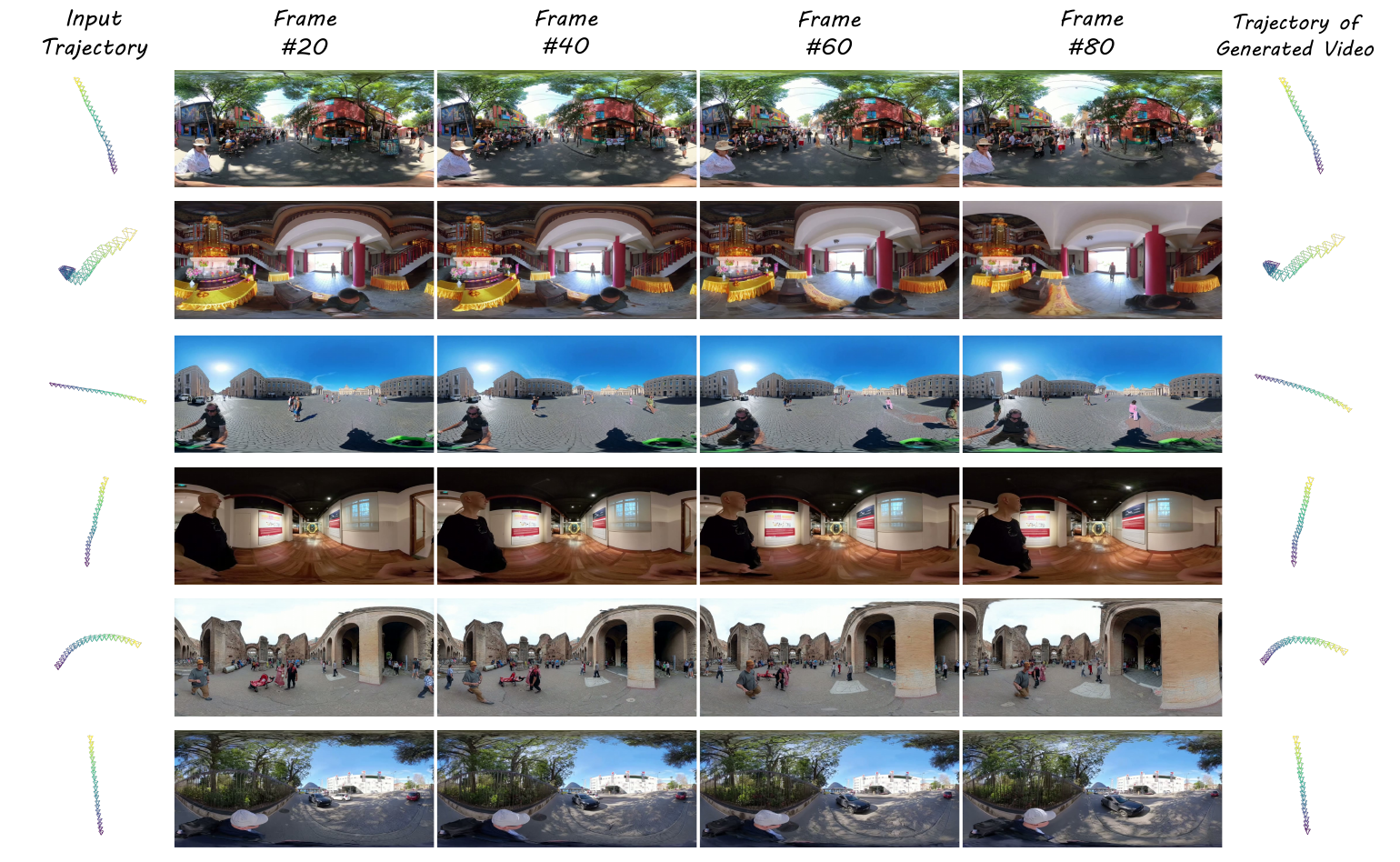}
  \caption{Qualitative examples of videos generated by Wan360. Wan360 is a camera-controllable interactive panoramic video generation model towards world exploration. To show the ability of following camera pose, we annotate the trajectory of generated videos.}
  \Description{A six-row grid of panoramic video examples. Each row places an input camera trajectory on the left, generated frames 20, 40, 60, and 80 in the center, and the recovered trajectory on the right. The scenes include an outdoor market, an event hall, a plaza, an indoor corridor, a stone courtyard, and a tree-lined road. Across the rows, the recovered trajectories closely follow the input paths while scene content remains coherent over time.}
  \label{fig:demo}
\end{figure}

\begin{figure}[!htbp]
  \centering
  \includegraphics[width=0.90\linewidth]{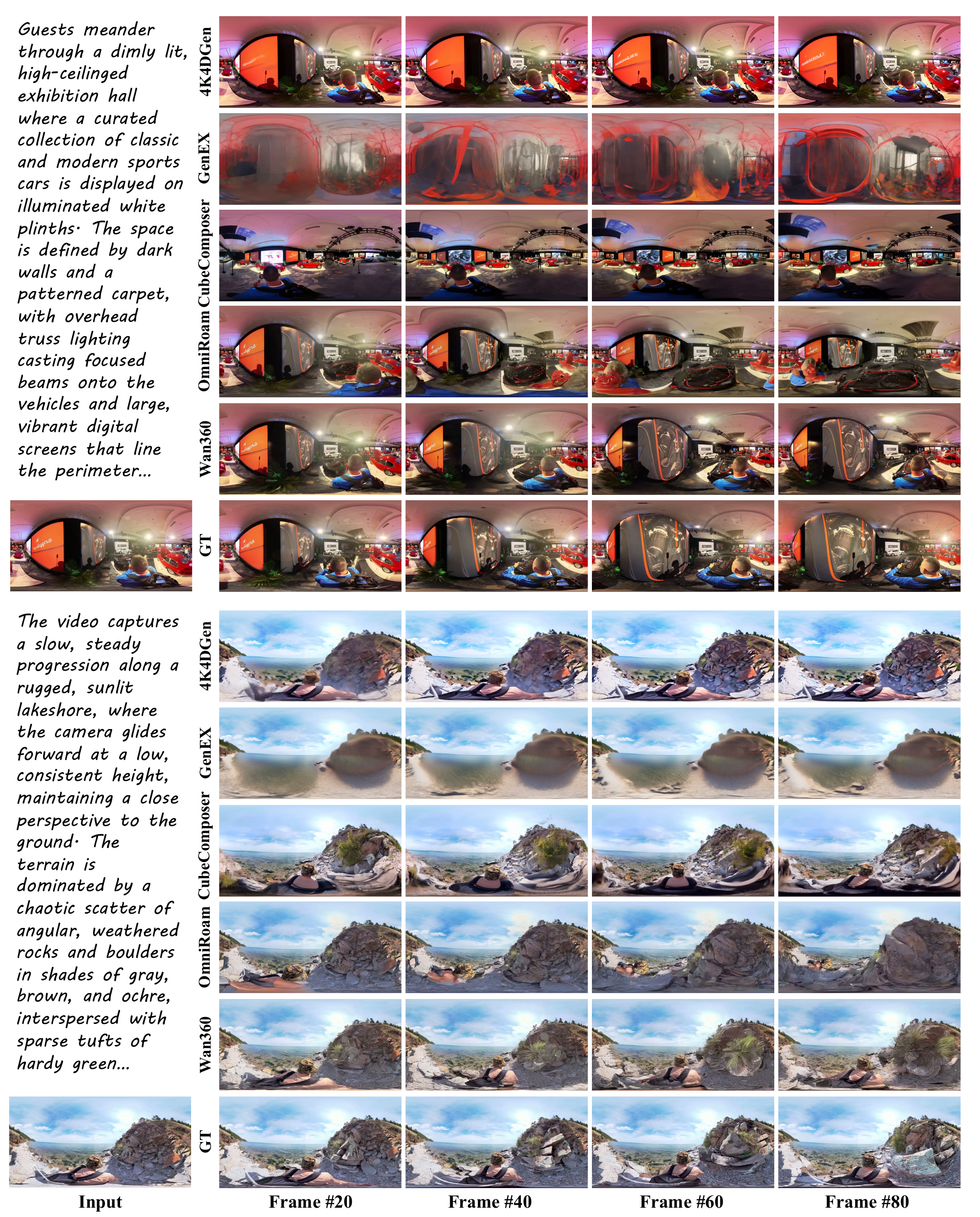}
  \caption{Qualitative comparison of panoramic video generation methods.  Wan360 produces panoramic videos with higher visual fidelity, better temporal consistency, and more realistic dynamics.}
  \Description{Two panoramic video examples compare 4K4DGen, GenHEX, CubeComposer, OmniRoam, Wan360, and ground truth at frames 20, 40, 60, and 80. The first example moves through a dim exhibition hall containing sports cars, and the second advances along a sunny rocky lakeshore. Wan360 preserves recognizable scene structure and produces temporal changes that most closely resemble the ground-truth sequences.}
  \label{fig:Qualitative}
\end{figure}

\cref{fig:Qualitative} provides qualitative comparisons with representative  methods.
4K4DGen~\cite{li20244k4dgenpanoramic4dgeneration} often produces limited apparent motion, GenEX~\cite{lu2025genexgeneratingexplorableworld} introduce artifacts and geometric distortions under large motion, CubeComposer~\cite{li2026cubecomposer} tends to produce less dynamic videos, and OmniRoam~\cite{liu2026omniroam} is constrained by static scene assumptions.
In contrast, Wan360 better preserves panoramic structure, visual details, and temporally coherent motion.

\begin{figure}[!htbp]
  \centering
  \includegraphics[width=0.99\linewidth]{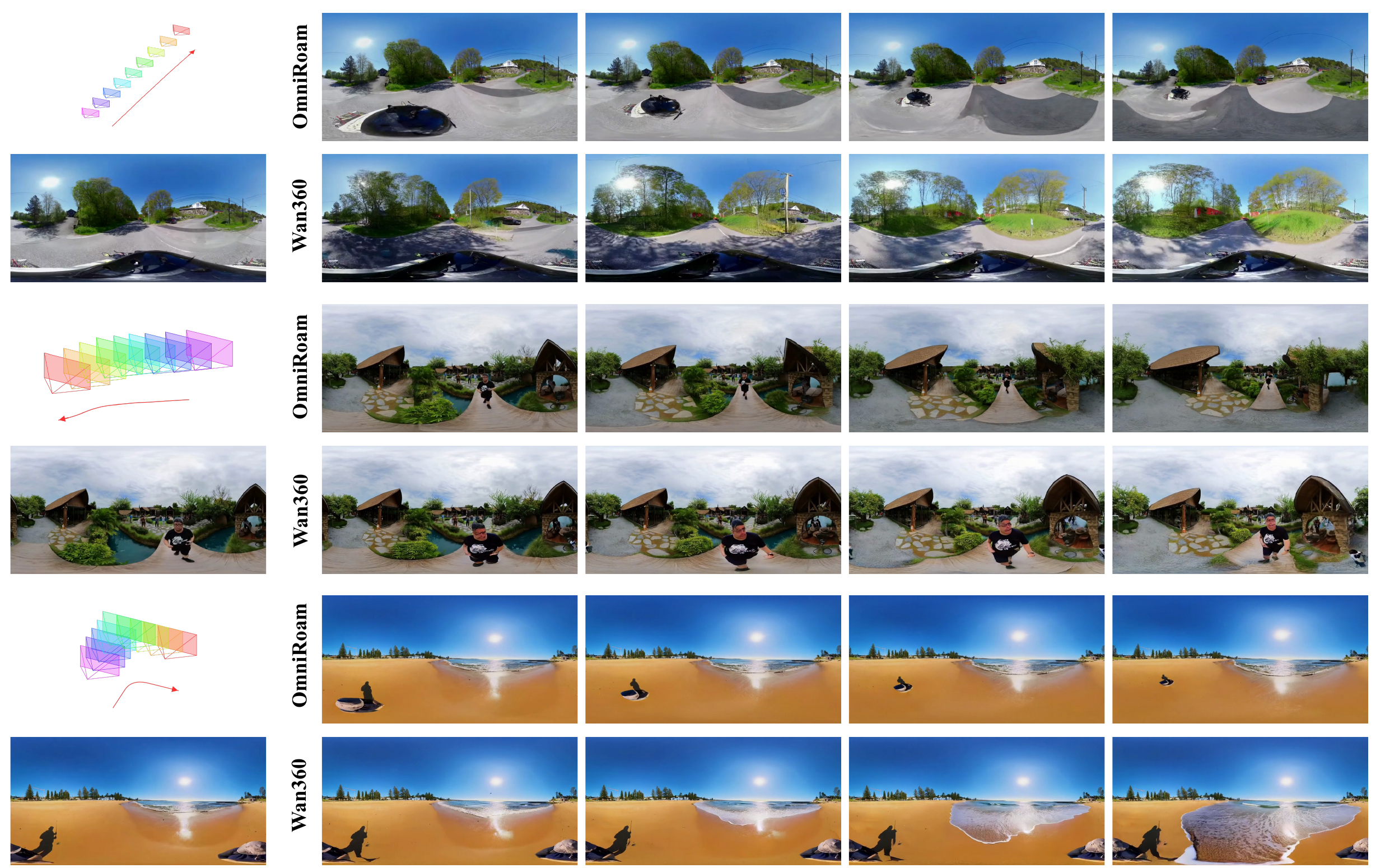}
  \caption{Qualitative comparison under target camera trajectories. The left column shows the input panorama and target trajectory, while the right columns compare generated sequences from OmniRoam~\cite{liu2026omniroam} and Wan360.}
  \Description{Three examples compare OmniRoam and Wan360 under specified camera paths. For each example, the left side shows a colored target trajectory and an input panorama, while four generated frames appear to the right. The scenes are a rural road, a garden walkway, and a beach. The Wan360 rows follow the requested translation and turning directions while retaining more stable panoramic scene structure.}
  \label{fig:pose_compare}
\end{figure}

\cref{fig:pose_compare} further compares Wan360 with OmniRoam under explicit target trajectories, showing that Wan360 follows the prescribed motion while maintaining dynamic real-world content.

\begin{figure}[!htbp]
  \centering
  \includegraphics[width=0.99\linewidth]{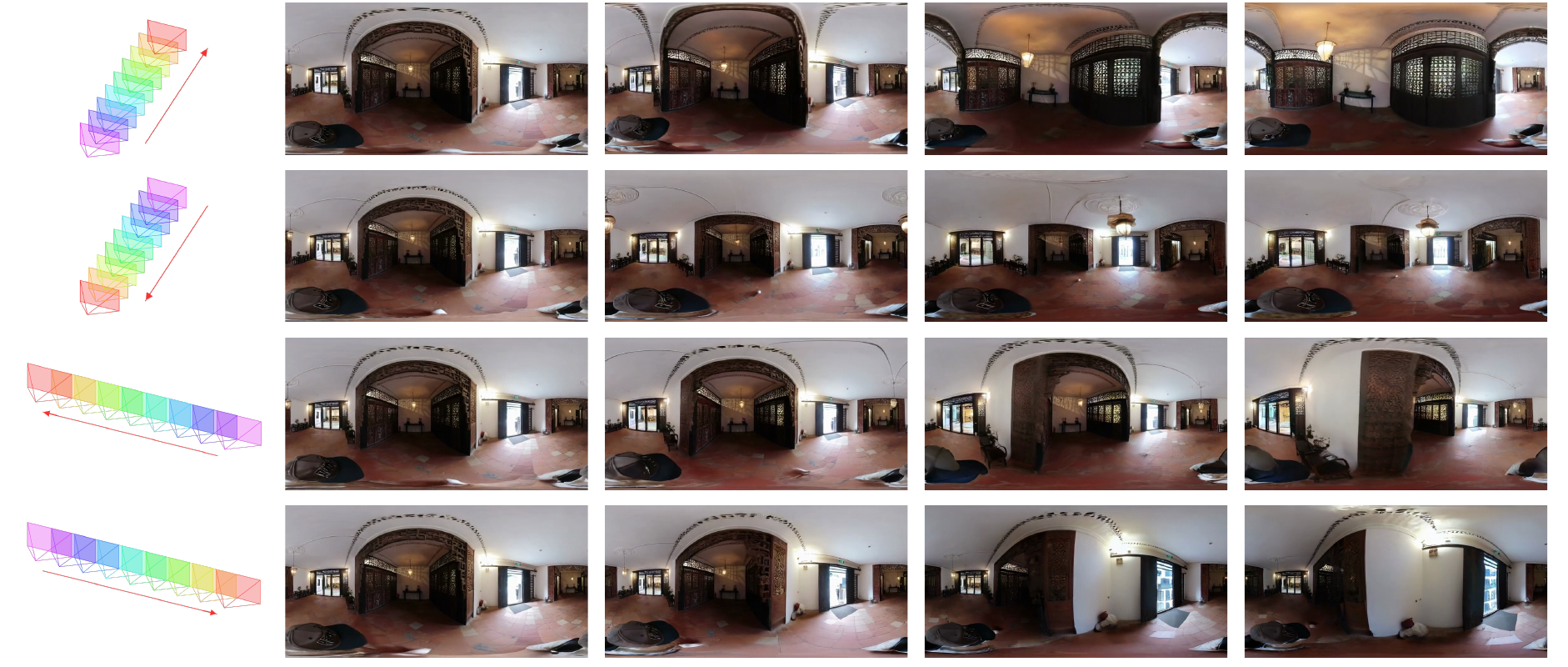}
  \caption{Qualitative results of Wan360 under different target trajectories for 10s. Given the same input panorama, Wan360 produces distinct video sequences aligned with different camera motions while maintaining consistent scene appearance.}
  \Description{Four rows start from the same panoramic view of an indoor hall with wooden partitions. A colored camera-path diagram at the left of each row specifies a different forward, backward, or lateral trajectory, and four generated frames show the corresponding viewpoint changes. The hall layout and appearance remain consistent across the distinct motions.}
  \label{fig:diff_pose}
\end{figure}

Finally, \cref{fig:diff_pose} shows examples generated from the same input panorama under different camera trajectories, demonstrating that Wan360 can produce trajectory-dependent panoramic videos with consistent scene appearance.

%% file: sec/6_conclusion.tex
\section{Conclusion}

In this work, we take a step toward interactive panoramic world exploration, where the central goal is to generate immersive 360\textdegree{} videos that remain coherent while following user-specified camera trajectories in dynamic real-world scenes.
This task requires both suitable annotated data and a model that can respect panoramic geometry under camera control.
To this end, MUGEN provides large-scale, high-quality real-world panoramic videos with rich semantic and geometric annotations, including explicit camera trajectories for controllable generation.
Built on MUGEN, Wan360 fine-tunes a perspective camera-control video generation baseline for ERP panoramas through periodic longitude RoPE, ERP-aware padding, random roll yaw, and a panoramic Pl\"ucker embedding.
Experiments show that MUGEN provides effective data support for the task, and that Wan360 generates panoramic videos with strong visual quality, temporal coherence, and trajectory controllability.
The proposed MUGEN and Wan360 establish a practical foundation for camera-controlled interactive panoramic world exploration in real-world environments.

\subsection{Limitations}
We use ViPE~\cite{huang2025vipe} to perform 3D reconstruction on videos generated by Wan360. In static scenes, the reconstructed scene structure and relative object positions remain generally stable, while the estimated camera trajectories are smooth and continuous. However, strict 3D consistency is not guaranteed, particularly in dynamic scenes with independently moving objects. Explicitly enforcing geometric and temporal consistency remains an important direction for future work.
Wan360 currently generates fixed-length clips and does not yet support real-time or unbounded long-horizon generation, where error accumulation remains an open challenge. In addition, MUGEN’s data primarily comes from the real world. In future work, we will consider extending it to other domains, such as animation and games.

